\documentclass[10pt, conference]{IEEEtran}
\usepackage{cite}
\usepackage{amsmath,amssymb,amsfonts}
\usepackage{algorithmic}
\usepackage{graphicx}
\usepackage{textcomp}
\usepackage{xcolor}

\usepackage[hidelinks]{hyperref}
\usepackage{multirow}
\usepackage{subcaption}
\usepackage{array}
\usepackage{float}

\def\BibTeX{{\rm B\kern-.05em{\sc i\kern-.025em b}\kern-.08em
    T\kern-.1667em\lower.7ex\hbox{E}\kern-.125emX}}
\begin{document}

\title{Real-Time Semantic Segmentation of Aerial Images Using an Embedded U-Net: A Comparison of CPU, GPU, and FPGA Workflows}

\author{
    \IEEEauthorblockN{
        Julien Posso\IEEEauthorrefmark{1}, 
        Hugo Kieffer\IEEEauthorrefmark{2}\IEEEauthorrefmark{3}, 
        Nicolas Menga\IEEEauthorrefmark{2}\IEEEauthorrefmark{4}, 
        Omar Hlimi\IEEEauthorrefmark{2}, 
        Sébastien Tarris\IEEEauthorrefmark{2}\IEEEauthorrefmark{3}, \\
        Hubert Guerard\IEEEauthorrefmark{5},
        Guy Bois\IEEEauthorrefmark{1}\IEEEauthorrefmark{5},
        Matthieu Couderc\IEEEauthorrefmark{2}\IEEEauthorrefmark{4},
        Eric Jenn\IEEEauthorrefmark{2}
    }
    \IEEEauthorblockA{\IEEEauthorrefmark{1} École Polytechnique de Montréal}
    \IEEEauthorblockA{\IEEEauthorrefmark{2} IRT Saint Exupéry}
    \IEEEauthorblockA{\IEEEauthorrefmark{3} Viveris Technologies}
    \IEEEauthorblockA{\IEEEauthorrefmark{4} Airbus Defence and Space}
    \IEEEauthorblockA{\IEEEauthorrefmark{5} Space Codesign Systems}
}

\maketitle

\begin{abstract}

This study introduces a lightweight U-Net model optimized for real-time semantic segmentation of aerial images, targeting the efficient utilization of Commercial Off-The-Shelf (COTS) embedded computing platforms. We maintain the accuracy of the U-Net on a real-world dataset while significantly reducing the model's parameters and Multiply-Accumulate (MAC) operations by a factor of 16. Our comprehensive analysis covers three hardware platforms (CPU, GPU, and FPGA) and five different toolchains (TVM, FINN, Vitis AI, TensorFlow GPU, and cuDNN), assessing each on metrics such as latency, power consumption, memory footprint, energy efficiency, and FPGA resource usage. The results highlight the trade-offs between these platforms and toolchains, with a particular focus on the practical deployment challenges in real-world applications. Our findings demonstrate that while the FPGA with Vitis AI emerges as the superior choice due to its performance, energy efficiency, and maturity, it requires specialized hardware knowledge, emphasizing the need for a balanced approach in selecting embedded computing solutions for semantic segmentation tasks.

\end{abstract}


\section{Introduction}

The advent of deep neural networks, especially Convolutional Neural Networks (CNNs), has revolutionized computer vision \cite{lecun_deep_2015}, introducing advanced capabilities for embedded systems in areas such as autonomous navigation \cite{sharma_pose_2018} and earth observation \cite{muhammad_efficient_2019, dimitrovski_current_2023, lv_land_2022}. Efficient hardware acceleration is vital for leveraging this technology, involving CPUs, GPUs, ASICs, FPGAs \cite{reuther_survey_2020}, and neural network compilers that bridge the gap between high-level Python libraries and hardware accelerators \cite{chen_tvm_2018}. These topics have recently gained significant attention, as discussed in Section~\ref{sec:related_works}. However, prior research has predominantly focused on image classification networks, specific hardware platforms, and compilers.

In this article, we present a pioneering, comprehensive, transversal study on the optimized implementation of image segmentation tasks for UAVs (Unmanned Aerial Vehicles) and satellites: specifically, the semantic segmentation of aerial images. We have enhanced a U-Net model for improved embeddability, reducing its parameters and MAC (Multiply-Accumulate) operations by a factor of 16 while maintaining accuracy. We evaluate and compare five implementation schemes (workflows) across three COTS (Commercial Off-The-Shelf) embedded computing platforms (GPU, CPU, FPGA), assessing them using metrics such as IoU (Intersection over Union), accuracy, power, throughput, energy efficiency, and memory footprint. We also consider engineering metrics like workflow maturity, usability, documentation, and community support. This study addresses key practical challenges and provides valuable insights for those looking to integrate deep neural networks into real-world applications.

The structure of this paper is organized as follows: Section~\ref{sec:related_works} reviews the literature pertinent to our research, providing foundational context. Section~\ref{sec:unet} details our computer vision task, specifically focusing on the semantic segmentation of aerial images using a lightweight U-Net to enhance its suitability for embedded systems. Section~\ref{sec:workflows} discusses the embedded computing platforms and examines the five workflows employed for implementing the neural network on these platforms. Section~\ref{sec:synthesis} synthesizes the main results, compares the workflows, and discusses the limitations of our study. Finally, Section~\ref{sec:conclusion} summarizes the study, highlighting the effectiveness of the workflows and the suitability of the hardware selections for our specific application domain.

\section{Related works}
\label{sec:related_works}

The quest for hardware accelerators is crucial for enabling real-time neural network inference. Central to this acceleration are technologies such as CPUs, GPUs, ASICs, and FPGAs \cite{reuther_survey_2020}. The role of compilers in bridging the gap between hardware capabilities and neural network performance is well-documented \cite{chen_tvm_2018}. Additionally, there is a noticeable shift in the embedded sector towards the adoption of Commercial Off-The-Shelf (COTS) computers \cite{perez_run-time_2020}.

Zhao et al. \cite{zhao_hardware_2018} and Li et al. \cite{li_deep_2020} meticulously review prevalent neural network compilers, including TVM, focusing primarily on their optimization mechanisms and their impact on the speedup of state-of-the-art image classification networks. Xing et al. \cite{xing_-depth_2019} provide an in-depth analysis of throughput, energy efficiency, and user-friendliness of six compilers, including TVM, aligning closely with our research. However, their analysis is confined to image classification networks such as ResNet50 and SqueezeNet, and they overlook potential quality degradation in neural network output due to the compilation and optimization processes.

Mittal et al. \cite{mittal_survey_2019} provide a detailed survey of Nvidia Jetson GPUs within the context of embedded systems, including their application in semantic segmentation networks. Abdelouahab et al. \cite{abdelouahab_accelerating_2018} and Guo et al. \cite{guo_dl_2019} review designs for neural network accelerators, with a particular emphasis on enhancing FPGA inference within image classification networks. Reuther et al. \cite{reuther_survey_2020} offer a comprehensive yet succinct survey of machine learning accelerators, focusing on performance and energy efficiency. Peccerillo et al. \cite{peccerillo_survey_2022} examine approximately 100 accelerators, exploring their diverse workflows.

Comparative studies on FPGA and GPU inference performance and energy efficiency for standard image classification networks are detailed by Nurvitadhi et al. \cite{nurvitadhi_can_2017}. Feng \cite{feng_computer_2019} compares FPGA and GPU inference, focusing solely on semantic segmentation networks on GPUs, notably excluding FPGAs. Li et al. \cite{li_gpu-outperforming_2018} highlight a performance comparison between FPGA and GPU inferences of binarized neural networks, revealing a trade-off between throughput and energy efficiency.

In the embedded domain, Dimitrovski et al. \cite{dimitrovski_current_2023} review neural network architectures for aerial imagery, primarily focusing on image classification accuracy while neglecting real-time inference capabilities. Wang et al. \cite{wang_aerial-bisenet_2021} and Wu et al. \cite{wu_towards_2019} propose new neural network architectures for real-time semantic segmentation of aerial images, yet their deployment on embedded hardware remains unexplored. Moreover, existing research often limits its focus to single COTS platforms and toolchains for real-time inference \cite{tijtgat_embedded_2017, mittal_survey_2019}.

The literature exhibits significant limitations, predominantly focusing on image classification networks, which are less relevant for earth observation via UAVs and satellites. Furthermore, the research largely relies on benchmark datasets (\textit{e.g.}, ImageNet) and often restricts its experimental scope to single COTS platforms and toolchains. Studies encompassing multiple hardware targets or compilers are typically classified as surveys rather than experimental research.

In contrast, our research stands out due to its comprehensive approach in several key areas:
\begin{itemize}
  \item A focus on semantic segmentation, an essential task for analyzing imagery from UAVs and satellites, diverging from the common focus on image classification.
  \item The adoption of a U-Net architecture for image segmentation, which includes both down-sampling (encoder) and up-sampling (decoder) paths, contrasting with the solely down-sampling nature of image classification networks. This approach exposes unique challenges in certain workflows that previous studies have not addressed.
  \item The utilization of the \textit{Inria Aerial Image Labeling Dataset} for real-world applications, moving away from the conventional use of benchmark datasets like ImageNet.
  \item A comprehensive evaluation involving multiple workflows and hardware targets, providing a holistic view of their performance and limitations.
\end{itemize}

\section{Embeddable U-Net-based Semantic Segmentation of Aerial Images}
\label{sec:unet}

\subsection{Semantic Segmentation of Aerial Images}
\label{sec:task}

Our research is situated within the context of earth observation, focusing primarily on two application domains: satellites and UAVs. These platforms are pivotal in acquiring high-resolution terrestrial imagery, offering spatial resolutions ranging from 0.2 to 10 meters, which are critical for numerous remote sensing applications \cite{muhammad_efficient_2019, dimitrovski_current_2023, lv_land_2022}. The primary limitation lies in the downlink capacity, as satellites and UAVs lack the capability to transmit all captured images to ground stations. Consequently, on-board analysis becomes essential to ensure that only relevant data is transmitted to Earth, optimizing both bandwidth and data relevance \cite{karapetyan_satellite_2015}.

In this context, semantic segmentation is indispensable as it enables precise on-board analysis of the high-resolution imagery acquired by satellites and UAVs. We employ the \textit{Inria Aerial Image Labeling Dataset} provided by Inria, renowned for its utility in benchmarking the generalization capabilities of semantic segmentation methodologies \cite{maggiori_can_2017}. This dataset includes 180 colored satellite photographs, each measuring 5000x5000 pixels (25 Megapixels). The primary task of the dataset involves semantic segmentation, which entails classifying each pixel of an input image into a specific category; in our case, this means distinguishing every pixel as either 'building' or 'not building'. This classification results in a segmentation map. Figure \ref{fig:pred_quality} illustrates this process. To optimize for training and model embeddability, we dissect these images into smaller segments of 256x256 pixels, maintaining slight overlaps. These segments are subsequently merged to reconstruct the original 5000x5000 segmentation map post-inference.

\subsection{U-Net Architecture}
\label{sec:unet_archi}

We selected a U-Net architecture for our workflow comparison. The U-Net \cite{ronneberger_u-net_2015}, initially proposed for biomedical image segmentation, has since become a widespread neural network architecture. It features a low number of parameters, a small memory footprint, and fewer MAC operations compared to other semantic segmentation networks, while still maintaining high accuracy. Additionally, it is designed to be trained with a limited amount of data, a common scenario in the embedded domain. These characteristics make the U-Net an ideal candidate for an embedded neural network.

\begin{figure}[h]
    \centering
    \includegraphics[scale=0.15]{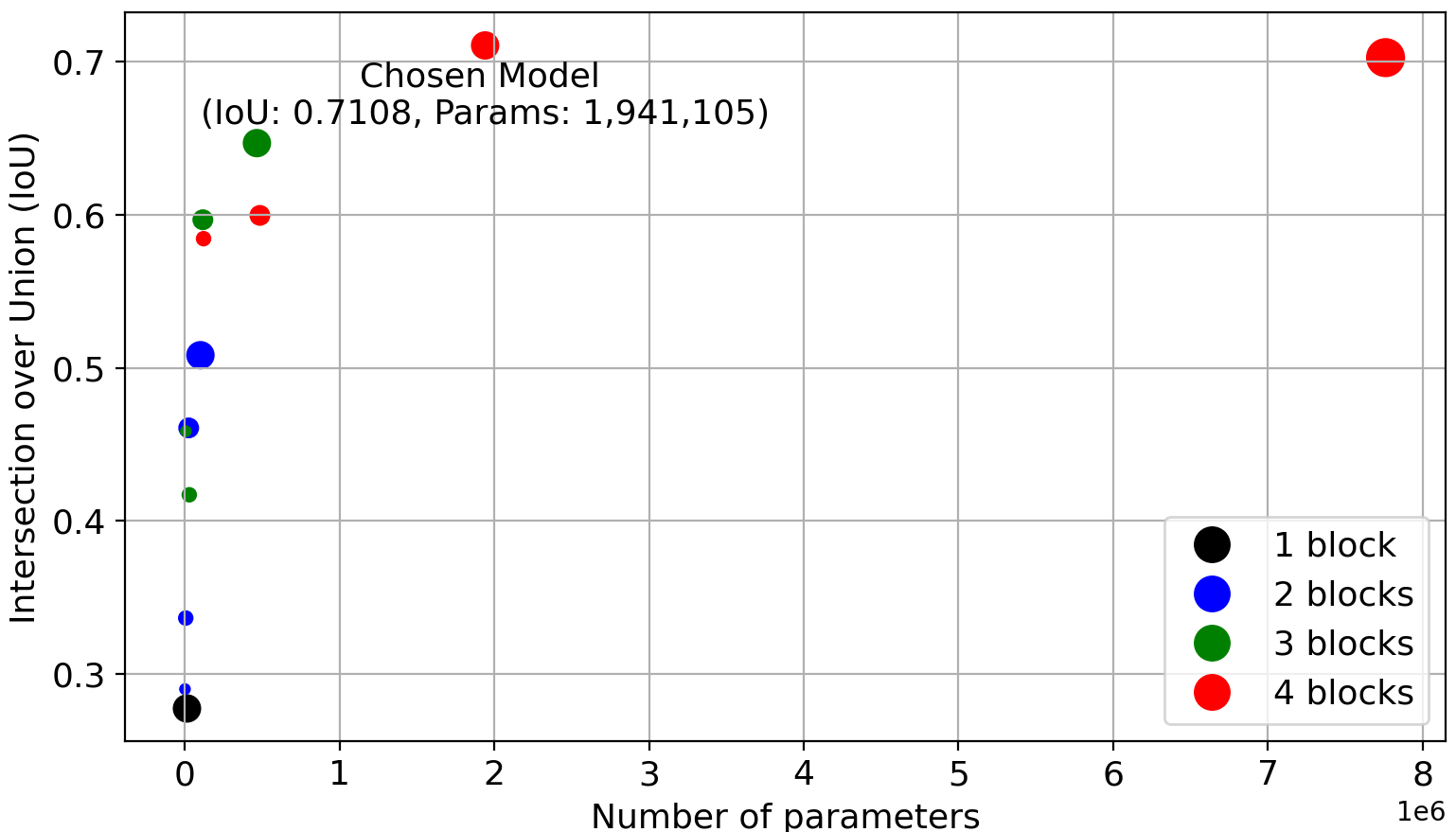}
    \caption{IoU on the validation set vs. the number of parameters of the U-Net. Circle size represents the number of channels.}
    \label{fig:trade_off_unet_size}
\end{figure}

However, we modified the U-Net to enhance its embeddability. We trained multiple versions of the U-Net, varying the number of layers and channels per layer. Figure \ref{fig:trade_off_unet_size} demonstrates the necessity of this process in an embedded context. In the down-sampling path of the U-Net, each block contains two convolutional layers and one max pooling layer. Similarly, in the up-sampling path, each block includes one transposed convolution and two convolutional layers. We adjusted the number of channels on each layer from $1/32$ to $1/2$ of the original U-Net and varied the number of blocks (\textit{i.e. the number of layers}) from one to four, while maintaining symmetry between the down-sampling and up-sampling paths of the U-Net.

\begin{figure*}[ht]
    \centering
    \includegraphics[scale=0.24]{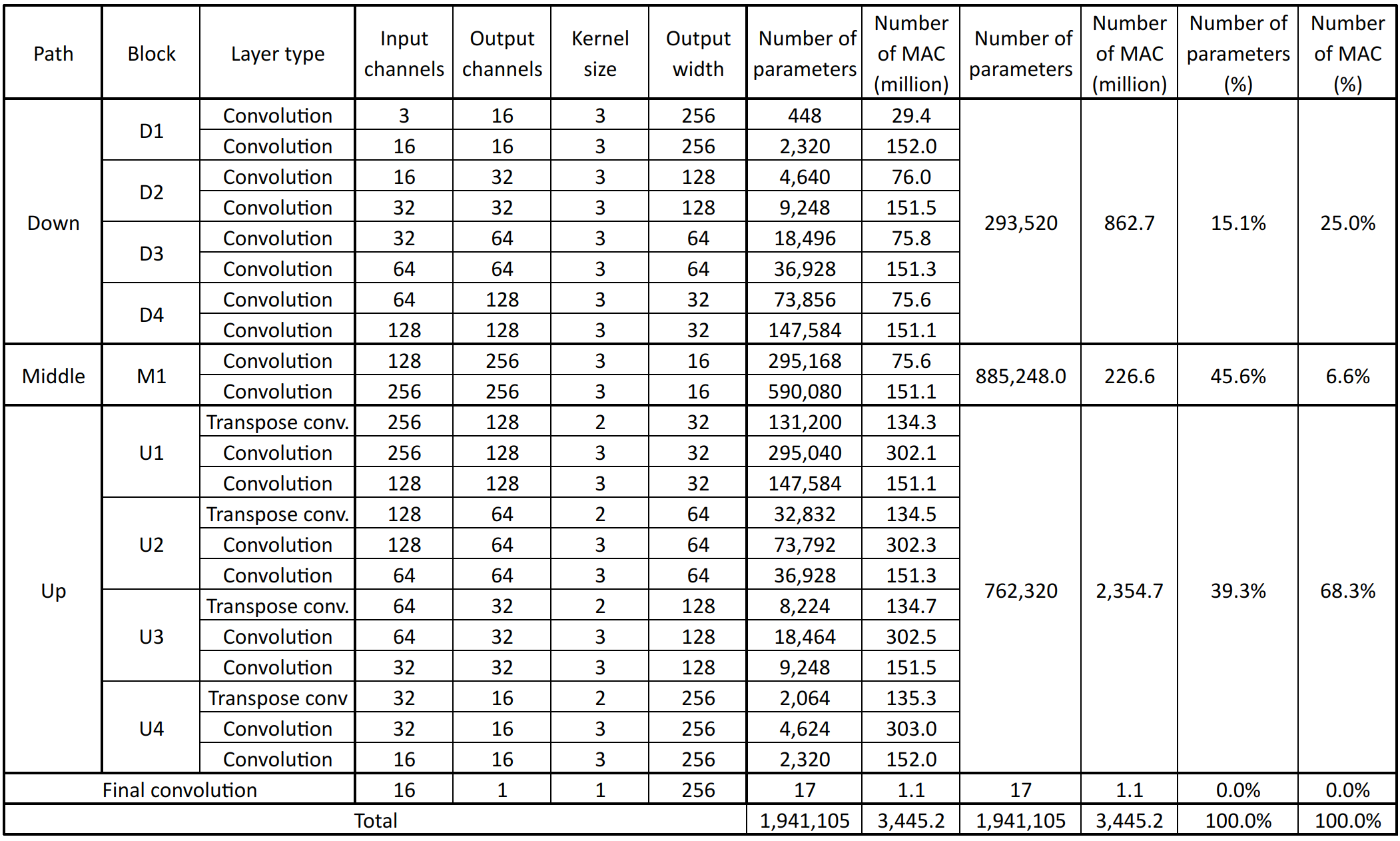}
    \caption{Detailed architecture of the U-Net model}
    \label{fig:unet_architecture}
\end{figure*}

We preserved the core structure of the original U-Net, which consists of four blocks, but reduced the number of channels per layer to one-fourth of the original. This adjustment significantly decreased the number of parameters (from 31 million to 1.9 million) and MAC (Multiply-Accumulate) operations (from 55 billion to 3.4 billion) required to process a single 256x256 image, while still maintaining accuracy on the \textit{Inria Aerial Image Labeling Dataset}. The number of parameters and MAC operations is proportional to the square of the number of channels, underscoring the importance of adapting neural network architectures to new datasets, especially in embedded contexts. Figure \ref{fig:unet_architecture} provides a detailed view of the U-Net architecture, showing the distribution of MAC operations and the number of parameters across the down-sampling (encoder), middle, and up-sampling (decoder) paths. Notably, the two middle layers of the U-Net contain almost half of the parameters, while the majority of MAC operations occur in the up-sampling path. This path is crucial for reconstructing the feature maps back to the original image size, explaining the higher number of MAC operations required for accurately generating the output segmentation map. The inclusion of transposed convolutions in the up-sampling path, not present in state-of-the-art image classification neural networks, introduces unique challenges in some workflows.

\subsection{Training Details}
\label{sec:training}

We trained our U-Net on an Nvidia RTX 3070 GPU, using Keras and TensorFlow 2.6, on the \textit{Inria Aerial Image Labeling Dataset}, as detailed in Section~\ref{sec:task}. Training began with random initial Float32 weights and utilized the Adam optimization algorithm \cite{kingma_adam_2017} with TensorFlow's default parameters and a learning rate of $1.0 \times 10^{-4}$ over 108 epochs. Training was halted after 15 epochs without improvement in the Intersection over Union (IoU) computed on the validation set. We employed the Binary Cross Entropy (BCE) loss function, which is effective for binary segmentation tasks. To enhance the model's robustness and reduce sensitivity to overfitting, we normalized the input images to a range between 0 and 1 and applied data augmentation techniques using OpenCV 2.5. These techniques included random rotations (multiples of 90 degrees) and horizontal and vertical flipping.

\subsection{U-Net Evaluation}

Table \ref{tab:eval_train} presents the evaluation of our lightweight U-Net, compared with the same data, task, and evaluation metrics used by the Inria team \cite{maggiori_can_2017}: the IoU of the building class and pixel accuracy. The Inria team employed a FCN (Fully Convolutional Network) followed by a MLP (Multi-Layer Perceptron). Additionally, they discuss the general training process but lack in-depth technical specifics about the architecture configurations, such as the number of parameters and layers. Nevertheless, the evaluations demonstrate that our lightweight U-Net outperforms the Inria team's neural network. The lightweight U-Net serves as a baseline for evaluating the five workflows explored in this paper.

\begin{table}[h]
    \centering
    \caption{Evaluation metrics of our lightweight U-Net on the validation set}
    \begin{tabular}{|l|c|c|}
        \hline
        Model & IoU & Accuracy \\
        \hline
        Lightweight U-Net (ours) & 0.7108 & 0.9546 \\
        \hline
        FCN + MLP (Inria) \cite{maggiori_can_2017} & 0.6467 & 0.9442 \\
        \hline
    \end{tabular}
    \label{tab:eval_train}
\end{table}

\begin{figure*}[ht]
     \centering
     \begin{subfigure}[b]{0.30\textwidth}
         \centering
         \includegraphics[width=\textwidth]{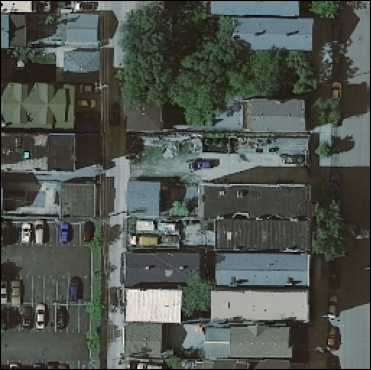}
         \caption{Input image}
         \label{fig:input_image}
     \end{subfigure}
     \hfill
     \begin{subfigure}[b]{0.30\textwidth}
         \centering
         \includegraphics[width=\textwidth]{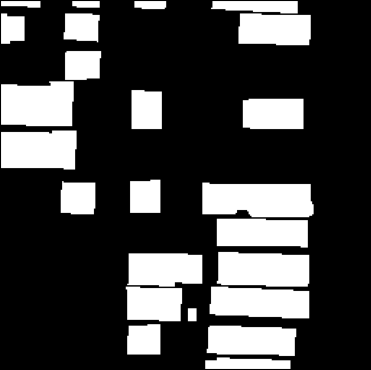}
         \caption{Ground truth}
         \label{fig:ground_truth}
     \end{subfigure}
     \hfill
     \begin{subfigure}[b]{0.30\textwidth}
         \centering
         \includegraphics[width=\textwidth]{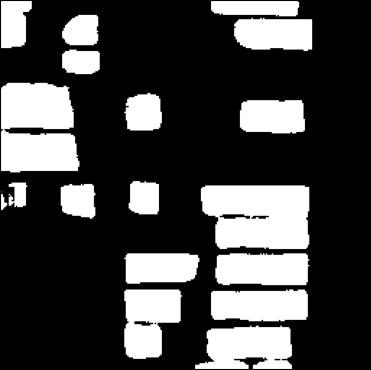}
         \caption{U-Net prediction}
         \label{fig:all_1x1_projection_weights}
     \end{subfigure}
        \caption{Qualitative evaluation of our Float32 Keras lightweight U-Net on a 256x256 image of the validation set}
        \label{fig:pred_quality}
\end{figure*}

Figure \ref{fig:pred_quality} shows an example of our U-Net's prediction quality compared to the ground truth on a 256x256 image. The buildings are generally well-predicted by the neural network, even if the contours of the predicted buildings are somewhat blurred, a similar effect was noticed in the original Inria publication \cite{maggiori_can_2017}.

\section{Platforms and Workflows}
\label{sec:workflows}

\subsection{Platforms for Real-Time Inference}

We selected two COTS platforms, specifically designed for embedded applications, to deploy our U-Net model. The Xilinx Zynq UltraScale+ MPSoC, equipped with four ARM Cortex-A53 processor cores and programmable logic (commonly referred to as an FPGA), has proven effective in both UAV \cite{kovari_mpdrone_2021} and space domains \cite{perez_run-time_2020}. For our implementation, we utilized three Xilinx Zynq UltraScale+ boards—Ultra96, ZCU102, and ZCU104—each equipped with the same processor but featuring varying FPGA sizes, to host the hardware accelerators. Nvidia Jetson platforms have also emerged as strong contenders for real-time inference of neural network-based vision algorithms, demonstrating applicability in UAV \cite{tijtgat_embedded_2017} and space domains \cite{adams_towards_2019}. Specifically, we employed the Nvidia Jetson AGX Xavier System on Module, which boasts eight ARM Cortex-A57 processor cores and an integrated GPU, enhancing the acceleration of neural network inference.

\subsection{Workflows Overview}

We evaluated various workflows to implement our U-Net on CPU, GPU, and FPGA platforms. On the GPU side, we first assessed the straightforward TensorFlow implementation, comparing it with the more complex but optimized Nvidia cuDNN library to understand the trade-offs between ease of use and performance. For the CPU, we utilized TVM, which is renowned for supporting major Python frameworks and offering the best speedup among neural network compilers \cite{chen_tvm_2018}, further enhanced by its auto-scheduling feature. For the FPGA, we explored both the open-source FINN framework and Xilinx's commercial DPU within the Vitis-AI toolchain. Although Vitis-AI is considered more mature, FINN offers experimental yet highly optimized options for creating optimized dataflow implementations \cite{blott_finn-_2018}. The following sections will delve into the details of each workflow.

\subsection{GPU Implementation with TensorFlow}

\subsubsection{Workflow Overview}

\begin{figure}[ht]
    \centering
    \includegraphics[scale=0.18]{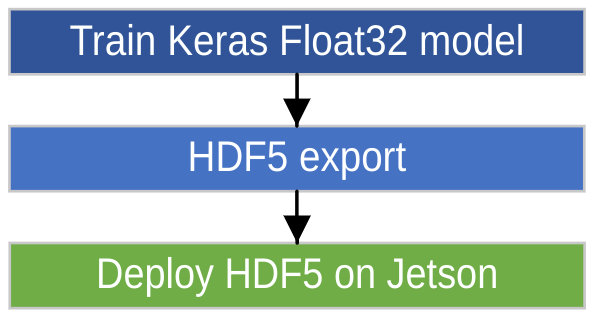}
    \caption{GPU workflow from Keras/TensorFlow training to Nvidia Jetson AGX Xavier inference using TensorFlow}
    \label{fig:gpu_flow_tf2}
\end{figure}

Figure \ref{fig:gpu_flow_tf2} presents the workflow used to deploy our model on the Nvidia Jetson AGX Xavier using TensorFlow 2.6. This workflow is straightforward, starting with the training of a Float32 model using Keras, serving as our baseline for evaluating GPU workflows. Notably, the model remains in Float32 format throughout, since quantization is only available in TensorFlow Lite. Our aim was to evaluate the most direct method for deploying a neural network on a Jetson GPU. Furthermore, the Jetson GPU efficiently processes Float32 operations on its CUDA (Compute Unified Device Architecture) cores. The trained model is exported in HDF5 format and then loaded onto the Nvidia Jetson AGX Xavier development kit. Onboard inference is conducted through a Python script, representing the simplest deployment method on the Nvidia Jetson platform, which operates on a Linux-based system with a Python stack, including TensorFlow.

\subsubsection{Quantitative Evaluation}

Table \ref{tab:eval_tf2} presents the evaluation metrics measured on the validation set throughout the TensorFlow workflow. The first row shows the results following training with Keras and TensorFlow in a Float32 format. Subsequent rows detail these metrics when the model is deployed on an Nvidia Jetson AGX Xavier board. The consistency observed between the standard computing environment and the embedded deployment is expected because the underlying model remains unchanged between the training and deployment stages.

\begin{table}[ht]
    \centering
    \caption{Evaluation metrics along the TensorFlow workflow}
    \begin{tabular}{|l|c|c|}
        \hline
        Model                 & IoU & Accuracy \\
        \hline
        Float32 Keras         & 0.7062 & 0.9594 \\
        Jetson implementation & 0.7062 & 0.9594 \\
        \hline
    \end{tabular}
    \label{tab:eval_tf2}
\end{table}

Table \ref{tab:tf2_board} summarizes the implementation metrics measured on the Jetson AGX Xavier. In this experiment, we varied the batch size to analyze its impact on the implementation metrics. Increasing the batch size to eight proved beneficial for improving throughput and energy efficiency while maintaining a reasonable memory footprint. The memory footprint includes the space needed for the model weights and activation functions, the batch of images, and additional Python libraries such as TensorFlow. Further increases in batch size did not yield significant benefits and resulted in an increased memory footprint, making a batch size of eight an optimal trade-off. A batch size of one is deemed beneficial only when memory footprint or latency is prioritized over throughput or energy efficiency. During the experiments, we noticed some variability in execution time, particularly for the first inference. The first inference with a batch of eight images took 238 milliseconds, while the subsequent inferences averaged around 107 milliseconds (plus or minus 10 milliseconds). The table also reports the average throughput for the entire validation set. The observed variability was consistent across all batch sizes, highlighting the importance of also considering the Worst Case Execution Time (WCET) in embedded systems where it is a critical factor.

\begin{table}[ht]
    \centering
    \caption{Implementation metrics on the Nvidia Jetson AGX Xavier with TensorFlow}
    \begin{tabular}{|m{0.7cm}|m{1.3cm}|m{1.0cm}|m{1.5cm}|m{1.3cm}|}
        \hline
        Batch size & Throughput (FPS) & Power (W) & Energy efficiency (mJ/image) & Memory (GB) \\
        \hline
        1 & 61.6 & 13.65 & 221.6 & 1.7 \\
        \hline
        8 & 74.6 & 14.56 & 195.2 & 2.2 \\
        \hline
        16 & 78.6 & 14.56 & 185.2 & 5.05 \\
        \hline
        32 & 75.8 & 14.56 & 192.1 & 5.3 \\
        \hline
    \end{tabular}
    \label{tab:tf2_board}
\end{table}

\subsubsection{Qualitative Evaluation}

The TensorFlow workflow targeting the Nvidia Jetson GPU is mature, straightforward, and well-documented, supported by an active community with numerous users, examples, and online tutorials. However, optimization of the neural network is limited within this framework. The high memory footprint presents significant concerns for embedded systems, which are often resource-limited compared to typical desktop or server environments. Furthermore, this high memory footprint could impact performance, energy efficiency, cost, and system stability, especially when the hardware is required to manage multiple applications simultaneously.

\subsection{GPU Implementation with CuDNN}

\subsubsection{Workflow Overview}

\begin{figure}[ht]
    \centering
    \includegraphics[scale=0.17]{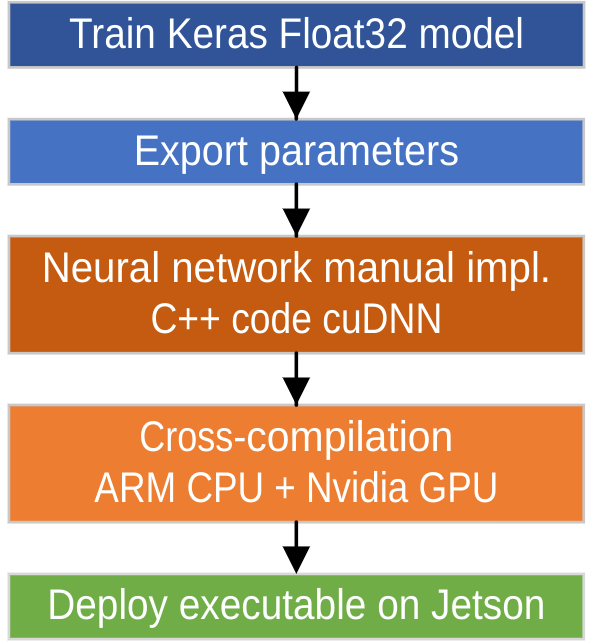}
    \caption{GPU workflow from Keras/TensorFlow training to Nvidia Jetson AGX Xavier inference with cuDNN}
    \label{fig:cudnn_workflow}
\end{figure}

Figure \ref{fig:cudnn_workflow} illustrates the workflow used to deploy our model on the Nvidia Jetson AGX Xavier utilizing the Nvidia cuDNN 8.4.1 library. Initially, we train a Float32 version of the model using Keras and export the trained parameters. Similar to the previous workflow, the model remains in Float32 format because quantization is only supported in TensorFlow Lite. The Jetson GPU is capable of efficiently processing Float32 operations on its CUDA cores. Subsequently, the neural network must be manually implemented in C++ with calls to the cuDNN library to execute operations on the GPU. The neural network is then cross-compiled for an ARM target using g++ and NVCC (Nvidia CUDA Compiler), resulting in an executable that is deployed on the Nvidia Jetson AGX Xavier, which operates a Linux-based system with the cuDNN library installed.

\subsubsection{Quantitative Evaluation}

During the evaluation, we encountered challenges, particularly due to the lack of a cuDNN implementation for the transposed convolution in the up-sampling path of the U-Net, as well as for the nearest neighbor upsampling operation. A feasible solution could have been to implement these layers in a custom CUDA program; however, due to limited time and inadequate support on the Nvidia forum, this approach was not viable. We successfully implemented the down-sampling path and the middle convolution of the U-Net using cuDNN. The implementation's accuracy was validated by comparing the intermediate tensor outputs from the middle convolution produced by cuDNN with those from TensorFlow, finding them equivalent within an absolute tolerance of $1e-8$. Thus, we conclude that the cuDNN workflow is unlikely to alter the evaluation metrics significantly.

Table \ref{tab:cudnn_board} presents the evaluation metrics measured on the validation set for the implemented down-sampling path and middle convolution of the U-Net using cuDNN. We estimated the full U-Net implementation performance by considering that the down-sampling path and middle convolutions comprise 31.6\% of the MAC operations, and we scaled the measured latency accordingly to estimate the total latency. Similarly, since these components represent 60.7\% of the parameters and intermediate feature maps, we adjusted the memory footprint to estimate the total memory usage. These estimates should be interpreted with caution.

\begin{table}[ht]
    \centering
    \caption{Measured and estimated implementation metrics on the Nvidia Jetson AGX Xavier with cuDNN}
    \begin{tabular}{|m{1.5cm}|m{1.1cm}|m{0.8cm}|m{1.3cm}|m{1.1cm}|}
        \hline
        Model & Latency (ms) & Power (W) & Energy efficiency (mJ/image) & Memory (MB) \\
        \hline
        Partial U-Net (measured) & 5.82 & 5.61 & 32.6 & 795 \\
        \hline
        U-Net (estimated) & 18.4 & 5.61 & 103.3 & 1310 \\
        \hline
    \end{tabular}
    \label{tab:cudnn_board}
\end{table}

\subsubsection{Qualitative Evaluation}

The cuDNN workflow for targeting Nvidia-embedded GPUs is mature yet intricate. cuDNN is primarily designed for developers of deep neural network (DNN) frameworks such as PyTorch or TensorFlow \cite{brown_accelerate_2014}. Consequently, it is more complex than other libraries and lacks extensive examples. Additionally, the absence of certain neural network layers necessitates a proficiency in CUDA programming, which is considerably more complex than using cuDNN alone. We also encountered discrepancies between the documentation and the actual implementation, which compounded the difficulty. The level of community activity is low; for instance, some queries on the Nvidia forums, particularly concerning transposed convolutions, have remained unanswered for over a year. While cuDNN is the optimal choice for achieving an optimized GPU implementation, especially where the memory footprint is a concern, this advantage requires a significantly greater development effort, particularly for neural networks that include layers not supported by the library.

\subsection{CPU Implementation with TVM}

\subsubsection{Workflow Overview}
\label{sec:tvm_overview}

\begin{figure}[ht]
    \centering
    \includegraphics[scale=0.17]{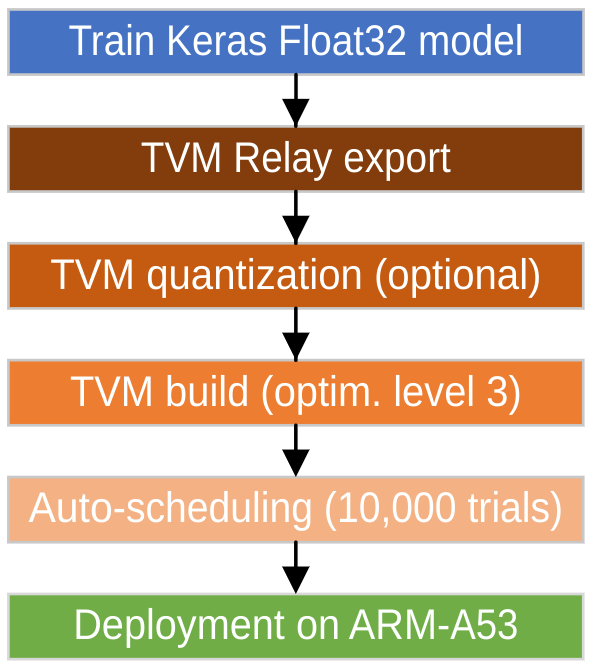}
    \caption{CPU workflow from Keras/TensorFlow training to ARM CPU inference with TVM}
    \label{fig:cpu_flow}
\end{figure}

Figure \ref{fig:cpu_flow} presents the workflow used to deploy our model on an ARM processor. We began by training a Float32 model with Keras, then utilized TVM 0.8 to export the model to Relay, TVM's intermediate graph representation. At this stage, quantization of the neural network is optional, which we discuss further in section \ref{sec:tvm_results}. We compiled the model using an optimization level of 3, which in our experiments achieved the best trade-off between optimization and neural network accuracy. Subsequently, we employed TVM's auto-scheduling, conducting 10,000 trials to optimize the scheduling of the inference on the CPU. The model was then ready for deployment on the ARM-A53 target, operating under a Linux-based system with the TVM runtime installed.

\subsubsection{Quantitative Evaluation}
\label{sec:tvm_results}

Table \ref{tab:eval_cpu} shows the evaluation metrics obtained on a subset of the validation set, consisting of 1500 images, used in the TVM workflow, as the full validation set execution time was prohibitively slow on board. To ensure consistency, we maintained the same sub-validation set from the Keras evaluation through to the onboard evaluation. The table initially reports the metrics following Float32 training with Keras and TensorFlow. Subsequent rows display the metrics obtained when deploying the neural network on an Ultra96 board. The TVM workflow, without quantization, preserved the quality of the neural network's output. In further experiments, we quantized every weight and activation function to eight bits, except for the first convolutional layer. We found that post-training quantization with TVM had a negligible impact on the evaluation metrics, minimally affecting both IoU and accuracy.

\begin{table}[ht]
    \centering  
    \caption{Evaluation metrics along the TVM workflow}
    \begin{tabular}{|l|c|c|}
        \hline
        Model                & IoU & Accuracy \\
        \hline
        Float32 Keras        & 0.7170 & 0.9546 \\
        Float32 TVM          & 0.7170 & 0.9546 \\
        Int8 TVM             & 0.7007 & 0.9518 \\
        \hline
    \end{tabular}
    \label{tab:eval_cpu}
\end{table}

Table \ref{tab:tvm_board} summarizes the performance metrics measured on the Ultra96 and ZCU104 boards. The ZCU104 demonstrated approximately ten percent faster execution than the Ultra96, attributable to its faster DDR memory. However, latency on both boards was significant, limiting real-time inference of semantic segmentation neural networks on these CPUs. Power consumption averaged 1.1W at thermal equilibrium, which is relatively low and was consistent across both boards and quantization levels, as the ARM cores were fully utilized under all conditions. Quantization increased the execution time threefold, possibly due to a bug in the version of TVM used, suggesting that the auto-scheduling functionality may not be fully compatible with the quantized version of our network. Energy efficiency was slightly better on the ZCU104, but the difference was minimal, except with the quantized version, which showed a significant increase. The memory footprint was reduced further with quantization.

\begin{table}[ht]
    \centering
    \caption{Implementation metrics on the Xilinx Zynq Ultrascale+ boards with TVM}
    \begin{tabular}{|m{0.9cm}|m{0.8cm}|m{1.1cm}|m{1.1cm}|m{1.2cm}|m{1.0cm}|}
        \hline
        Board & Quanti-zation & Latency (ms) & Power (W) & Energy efficiency (J/image) & Memory (MB) \\
        \hline
        Ultra96 & No  & 540.7 & 1.05 & 0.568 & 68.4 \\
        Ultra96 & Yes & 1687  & 1.05 & 1.77  & 39.5 \\
        ZCU104  & No  & 489.2 & 1.11 & 0.543 & 78.7 \\
        \hline
    \end{tabular}
    \label{tab:tvm_board}
\end{table}

\subsubsection{Qualitative Evaluation}

The TVM workflow for targeting ARM CPUs is well-established, yet it is not without limitations, particularly due to a quantization bug encountered during our evaluations. This issue can be circumvented by utilizing the quantization functionalities of Keras/TensorFlow. The workflow benefits from being user-friendly, supported by extensive documentation and numerous examples. The versatility of the TVM stack allows for deployment on any ARM CPU that operates a Linux-based system, including smartphones and Raspberry Pi devices. Switching the target CPU requires altering only a single line of Python code. The community behind TVM is highly active, annually hosting TVMCon, a conference that fosters collaboration between academia and industry on neural network compilation. TVM's fully automated build and auto-scheduling processes facilitate the deployment and optimization of state-of-the-art convolutional neural networks, rendering the TVM workflow exceptionally adaptable.

\subsection{FPGA Implementation with FINN}

\subsubsection{Workflow Overview}

\begin{figure}[ht]
    \centering
    \includegraphics[scale=0.11]{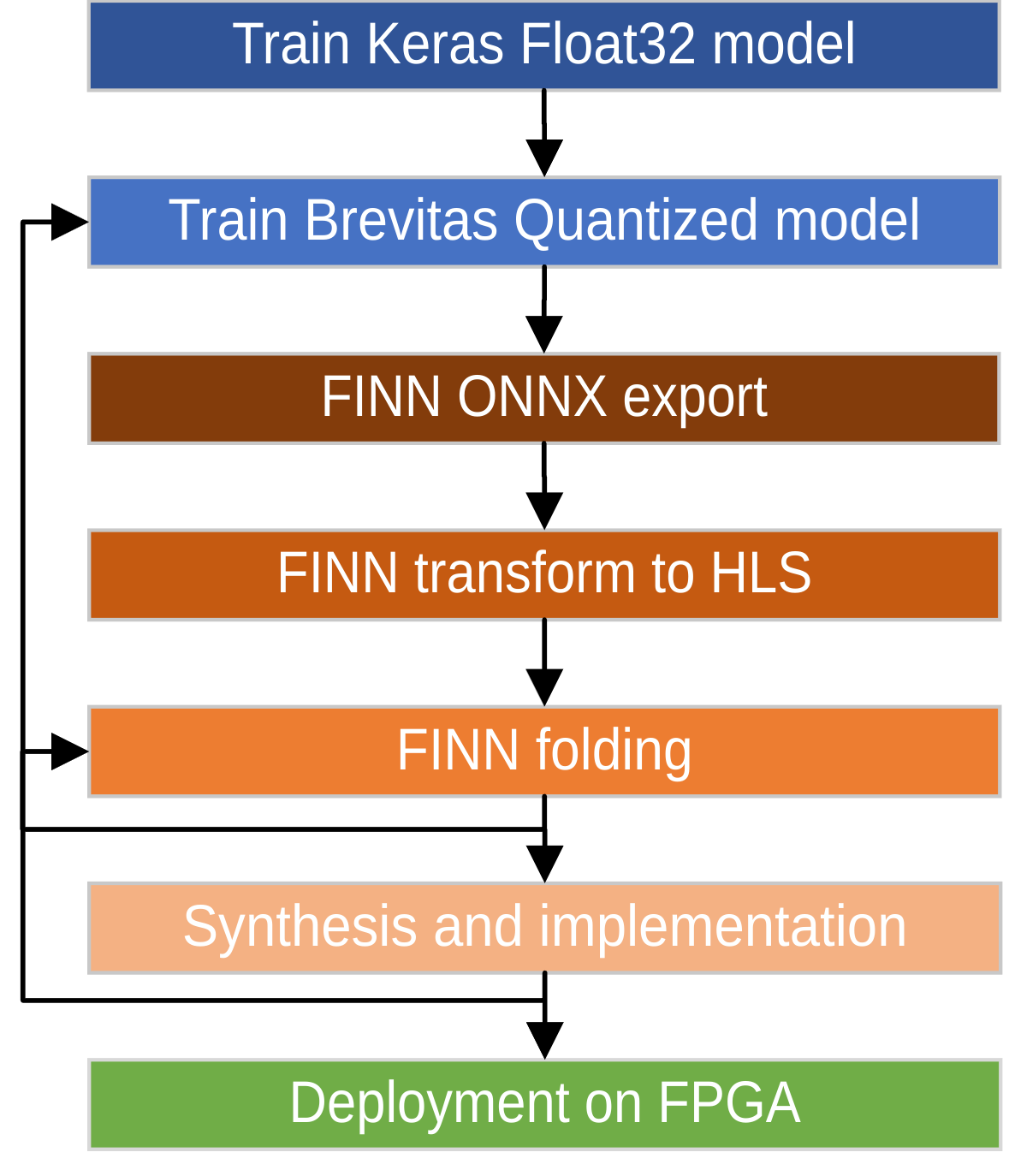}
    \caption{FPGA workflow from PyTorch/Brevitas training to FPGA inference using FINN}
    \label{fig:fpga_finn_flow}
\end{figure}

Figure \ref{fig:fpga_finn_flow} presents the workflow utilized to deploy our model on an FPGA using the FINN library. As FINN is incompatible with Keras or TensorFlow, we re-implemented the U-Net model using PyTorch 1.7.1 and Brevitas 0.6.1. Brevitas is a quantization library designed to facilitate Quantization Aware Training (QAT) with PyTorch and to support deployment through FINN \cite{pappalardo_xilinxbrevitas_2021}. Initially, we trained a Float32 version of the U-Net using Keras and exported the weights to the PyTorch/Brevitas version of the U-Net. We then proceeded with training a quantized version of the U-Net using QAT in PyTorch/Brevitas, starting from the Float32 weights to significantly reduce QAT duration. Brevitas supports mixed-precision quantization, enabling layer-wise bit-width parametrization for both weights and activation functions. After training, the model was exported to the ONNX format, which is compatible with FINN. At this stage, the model is transformed into a graph that contains only FINN HLS-compatible nodes. Subsequently, we defined the folding configuration for each graph node to set the parallelism, aiming to match the target latency without exceeding the FPGA’s available resources. If the folding configuration exceeded the FPGA resources, it required returning to the bit-width parametrization step and reiterating the QAT phase or adjusting the target latency. FINN's built-in functions facilitate the invocation of Vitis HLS to synthesize each node independently, integrate them, and then implement the combined solution as a Vivado 2022.1 project deployed on the FPGA. FINN also offers rapid prototyping capabilities using the Pynq library.

\subsubsection{Quantitative Evaluation}

Table \ref{tab:eval_finn} displays the evaluation metrics measured on the validation set throughout the FINN workflow. The initial row recalls the metrics after Float32 training with Keras. The final row presents the metrics for the quantized U-Net post-training, employing binary weights and 4-bit activation functions across all layers. Despite aggressive quantization, the accuracy and IoU only showed a slight decrease. Due to a suspected bug in the FINN library, we could not perform onboard inference to directly measure the evaluation metrics, a limitation we will discuss further in section \ref{sec:engineering_finn}.

\begin{table}[ht]
    \centering
    \caption{Evaluation metrics along the FINN workflow}
    \begin{tabular}{|l|c|c|}
        \hline
        Model                & IoU         & Accuracy \\
        \hline
        Float32 Keras        & 0.7108      & 0.9531 \\
        Quantized Brevitas   & 0.6837      & 0.9488 \\
        \hline
    \end{tabular}
    \label{tab:eval_finn}
\end{table}

While onboard inference execution was not possible, we derived certain results from the Vivado project, synthesis, and implementation reports. Table \ref{tab:finn_board} summarizes these findings and estimations. The latency was derived from the synthesis reports, considering the highest latency across all graph nodes (786,432 cycles) as the accelerator's initiation interval. With a clock frequency of 100 MHz, we estimated the accelerator's latency to be 7.86 milliseconds, corresponding to a throughput of 127 images per second. The on-chip power consumption, estimated at 5.5 Watts, was obtained from the FINN-generated Vivado project. The estimated energy efficiency is noteworthy, given the implementation of a low-bit quantized U-Net, although these results are provisional and should be approached with caution.

\begin{table}[ht]
    \centering
    \caption{Estimation of the implementation metrics on the Xilinx ZCU104 with the FINN workflow}
    \begin{tabular}{|m{1.0cm}|m{1.4cm}|m{1.0cm}|m{1.5cm}|m{1.3cm}|}
        \hline
        Board & Throughput (FPS) & Power (W) & Energy efficiency (J/image) & Memory (MB) \\
        \hline
        ZCU104 & 127.2 & 5.46 & 0.043 & N/A \\
        \hline
    \end{tabular}
    \label{tab:finn_board}
\end{table}

Table \ref{tab:fpga_resources_finn} provides a summary of FPGA resource utilization based on the post-implementation report generated by Vivado, highlighting LUTs (Lookup Tables) as the primary limiting factor. The LUTs are predominantly utilized for the convolution computations, namely the im2col algorithm and the matrix-vector multiplication unit. Notably, the multi-threshold layers, representing the quantized activation functions, also consume a substantial number of LUTs, proportional to the square of the bit-width of the activation functions. We chose binary weights and 4-bit activations as an optimal balance between accuracy and estimated throughput. This approach also eliminated the need for DSPs, reducing the resource demands significantly. Our experience has shown that the FPGA resource estimations provided by FINN's Python script were found to be unreliable.

\begin{table}[ht]
    \centering
    \caption{FINN FPGA resource usage on ZCU104 board}
    \begin{tabular}{|m{1.5cm}|m{1.7cm}|m{1.7cm}|m{1.5cm}|}
        \hline
        FPGA resource & Post-implementation utilization & FINN Python estimation & Available \\
        \hline
        LUT    & 205,249 (89\%) & 155,905 & 230,400 \\
        LUTRAM & 43,498 (43\%)  & Not Available &  101,760 \\
        Flip-Flop & 235,448 (51\%) & Not available & 460,800 \\
        BRAM & 96 (31\%) & 233 & 312 \\
        DSP & 0 (0\%) & 0 & 1,728 \\
        \hline
    \end{tabular}
    \label{tab:fpga_resources_finn}
\end{table}

\subsubsection{Qualitative Evaluation}
\label{sec:engineering_finn}

The Brevitas library for training quantized neural networks targeting FINN implementations is mature and user-friendly, closely mimicking the PyTorch experience, albeit lacking in examples. Conversely, the FINN library is still under development. We encountered and locally fixed several source code bugs during our experiments. While some of these issues have been addressed recently, indicating active development, the community remains relatively small compared to other libraries. The absence of certain HLS backend templates, such as transposed convolution, posed challenges. We circumvented this by substituting with a nearest neighbor upsampling layer followed by a convolution, which did not alter the U-Net’s parameter count or MAC operations.

Utilizing the FINN library can be challenging, particularly during the transformation phase, which requires users to meticulously determine the appropriate transformations and their sequence. Often, modifications to the network architecture and quantization scheme are necessary to remove non-HLS compatible nodes. We had to develop two custom transformations not present in FINN to synthesize the U-Net effectively. A significant issue related to the U-Net’s shortcuts prevented us from implementing the neural network on the FPGA. This issue could stem from a problem with our custom transformations, a bug in FINN's handling of concatenation layers, or FINN's algorithm not allocating sufficiently large FIFOs to store the activation functions of the down-sampling path, thereby hampering the up-sampling path’s ability to perform its convolutions. Additionally, the documentation, spread across various websites and GitHub pages, is fragmented and challenging to navigate.

The FINN library holds significant potential for energy-constrained applications and is poised to mature into a highly energy-efficient method for executing neural network inference on FPGAs. As it develops, FINN’s approach, with its capacity for mixed-precision quantization and configurable folding, will enable tailored optimization for each layer's bit-width, accuracy, resource usage, and latency.

\subsection{FPGA Implementation with Xilinx Vitis-AI}

\subsubsection{Workflow Overview}
\label{sec:vitis_overview}

\begin{figure}[ht]
    \centering
    \includegraphics[scale=0.12]{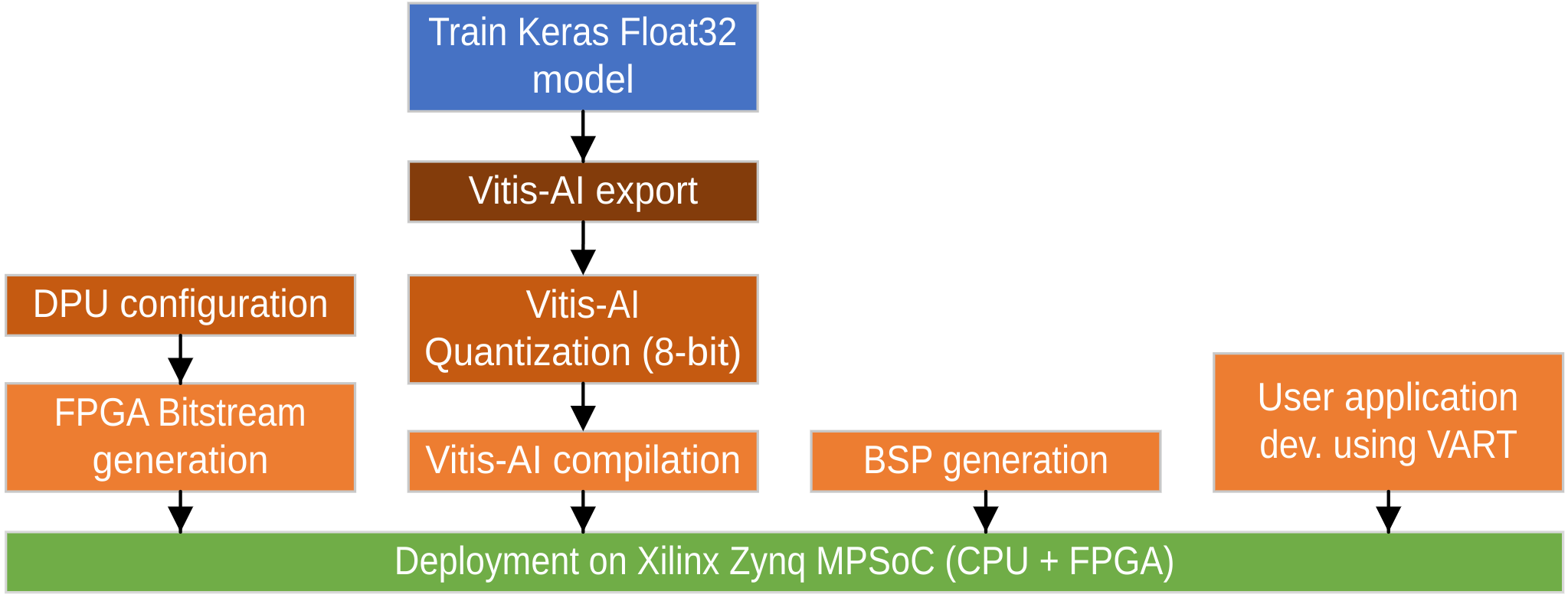}
    \caption{FPGA workflow from Keras/TensorFlow training to FPGA/CPU inference using Vitis-AI}
    \label{fig:fpga_vitis_flow}
\end{figure}

Figure \ref{fig:fpga_vitis_flow} outlines the workflow used to deploy a neural network on a Xilinx Zynq Ultrascale+ MPSoC using the Vitis-AI framework. This approach, distinct from rapid prototyping, is focused on actual embedded deployment. The process begins with training a Float32 model using Keras, followed by exporting it through the Vitis-AI toolkit version 2.0. Deployment on the MPSoC involves four primary activities:
\begin{itemize}
    \item Configuring the DPU (Deep Learning Processor Unit) accelerator and generating the FPGA bitstream. This includes selecting the number of DPU cores and their size, which dictates the operations per clock cycle.
    \item Generating the application code in C++ to orchestrate model execution using the VART (Vitis AI Runtime).
    \item Compiling the model using 8-bit quantization with the Vitis-AI tools.
    \item Creating the Board Support Package (BSP) for the ZCU102 board.
\end{itemize}

Following these steps, the model is executed on the MPSoC, with the DPU on the FPGA handling most of the network operations. However, the CPU may process some layers, particularly when specific functions like the sigmoid activation at the end are not supported by the Vitis-AI quantization.

\subsubsection{Quantitative Evaluation}

Table \ref{tab:eval_vitis} presents the evaluation metrics obtained from the validation set using the Vitis-AI workflow, with the initial line providing a baseline from Float32 training with Keras. Following the model's quantization to 8-bit using Vitis-AI, no loss in accuracy was observed, thanks to the toolkit's effective calibration function. The quantized model was subsequently deployed on the Xilinx DPU on the ZCU102 board, where no degradation in performance was noted, suggesting a possible regularization effect.

\begin{table}[ht]
    \centering
    \caption{Evaluation metrics along the Vitis-AI workflow}
    \begin{tabular}{|l|c|c|}
        \hline
        Model             & IoU         & Accuracy \\
        \hline
        Float32 Keras     & 0.7108      & 0.9531 \\
        Int8 Vitis        & 0.7156      & 0.9542 \\
        Int8 DPU          & 0.7263      & 0.9583 \\
        \hline
    \end{tabular}
    \label{tab:eval_vitis}
\end{table}

Table \ref{tab:vitis_board} summarizes the implementation metrics on the Xilinx ZCU102 board, measured on the validation set. The configuration uses three DPU cores, each capable of 4096 operations per clock cycle at 100 MHz. This setup was determined to be the best trade-off for embedded inference, balancing throughput and power consumption for optimal energy efficiency.

\begin{table}[ht]
    \centering
    \caption{Implementation metrics on the Xilinx ZCU102 with the Vitis-AI workflow}
    \begin{tabular}{|m{1.0cm}|m{1.4cm}|m{1.0cm}|m{1.5cm}|m{1.3cm}|}
        \hline
        Board & Throughput (FPS) & Power (W) & Energy efficiency (J/image) & Peak memory (MB) \\
        \hline
        ZCU102 & 46.9 & 2.51 & 53.5 & 31 \\
        \hline
    \end{tabular}
    \label{tab:vitis_board}
\end{table}

Table \ref{tab:fpga_resources_vitis} shows the FPGA resource utilization, with DSPs and BRAMs being the primary limiting factors due to their roles in MAC operations and storage of weights and intermediate feature maps, respectively. LUTs, LUTRAMs, and Flip-Flops still have available capacity, providing potential for future increases in the size or number of DPU cores.

\begin{table}[ht]
    \centering
    \caption{Vitis-AI FPGA resource usage with 3-core DPU on ZCU102 board}
    \begin{tabular}{|l|c|c|}
        \hline
        FPGA resource & Post-implementation utilization & Available \\
        \hline
        LUT    & 133,425 (49\%) & 274,080 \\
        LUTRAM & 17,027 (12\%)  & 144,000 \\
        Flip-Flop & 297,576 (54\%) & 548,160 \\
        BRAM & 771 (84\%) & 912 \\
        DSP & 2,070 (82\%) & 2520 \\
        \hline
    \end{tabular}
    \label{tab:fpga_resources_vitis}
\end{table}

\subsubsection{Qualitative Evaluation}

The Vitis-AI workflow is robust, demonstrating significant maturity, particularly with toolchain updates in versions 2.0 and 2.5 that resolved previously encountered bugs. This versatile workflow supports a wide array of neural network layers, and users can incorporate custom IP blocks to introduce new operations. Xilinx provides comprehensive documentation and end-to-end examples through the Vitis-AI Model Zoo. The community surrounding Vitis-AI has grown rapidly, although the learning curve remains steep due to the complexity of integrating various components such as BSP, Vivado, PetaLinux, and Vitis-AI tools. Additionally, while most components of Vitis AI are open source, some elements, such as the Vitis AI Compiler, remain proprietary, and certain tools within the Xilinx ecosystem require a commercial license.

\section{Synthesis}
\label{sec:synthesis}

\subsection{Synthesis and Workflow Comparison}
\label{sec:synthesis_comparison}

Table \ref{tab:synthesis_eval} synthesizes the evaluation and implementation metrics results across the five workflows. As discussed in Section \ref{sec:workflows}, onboard implementation was not achievable for the cuDNN and FINN workflows. Consequently, the implementation results from these workflows are estimates and should be interpreted with caution. Quantization is employed only when the hardware target does not support Float32 operations. The CPU and GPU workflows maintain the neural network’s output quality, thus achieving the same accuracy and Intersection over Union (IoU) as their respective baselines. The FINN workflow causes a slight degradation in accuracy and IoU, which is minimal considering the use of low-bit quantization. Conversely, the Vitis-AI workflow marginally improves the evaluation metrics on the validation set due to its quantization and calibration mechanisms, introducing a regularization effect. All workflows are compared at iso-accuracy levels. Nevertheless, there are significant differences in throughput and power consumption across the platforms and workflows. As expected, the CPU exhibits the lowest throughput, resulting in poor energy efficiency. The FPGA workflows, utilizing FINN or Vitis-AI, demonstrate superior energy efficiency. Both FINN and Vitis-AI enable the creation of customizable neural network accelerators, allowing for tailored FPGA resource usage, which in turn affects throughput and power consumption. Additionally, the use of quantization contributes to reduced power consumption. In contrast, GPU workflows and platforms have a considerably higher memory footprint compared to CPU and FPGA workflows and targets, presenting potential challenges in an embedded context.

\begin{table*}[ht]
    \centering
    \caption{Synthesis of the evaluation and implementation metrics of the five workflows}
    \label{tab:synthesis_eval}
    \begin{tabular}{|l|c|c|c|c|c|}
    \hline
        Platform & \multicolumn{2}{c|}{Nvidia GPU} & \multicolumn{3}{c|}{Xilinx Zynq UltraScale+ MPSoC} \\
        \hline
        Board & \multicolumn{2}{c|}{Jetson AGX Xavier} & \multicolumn{2}{c|}{ZCU104} & ZCU102 \\
        \hline
        Workflow & TensorFlow & cuDNN & TVM (CPU) & FINN (FPGA) & Vitis-AI (FPGA) \\
        \hline
        Implementation & Yes & No & Yes & No & Yes \\
        \hline
        Numeric precision & Float32 & Float32 & Float32 & W1A4 & Int8 \\
        \hline
        Accuracy change (vs. baseline) & 0\% & 0\% & 0\% & -0.43\% & +0.52\% \\
        \hline
        IoU change (vs. baseline) & 0 & 0 & 0 & -0.0271 & +0.0155 \\
        \hline
        Throughput (FPS) & 74.6 & 54.3 & 2.04 & 127 & 46.9 \\
        \hline
        Power (W) & 14.6 & 5.61 & 1.11 & 5.46 & 2.51 \\
        \hline
        Energy efficiency (mJ/image) & 195 & 103 & 543 & 43.0 & 53.5 \\
        \hline
        Memory (MB) & 2200 & 1310 & 78.70 & N/A & 31 \\
        \hline        
    \end{tabular}
\end{table*}

\begin{table*}[ht]
    \centering
    \caption{Synthesis of the engineering metrics of the five workflows. Metrics are quantified as high, medium, and low.}
    \label{tab:synthesis_engineering} 
    \begin{tabular}{|l|c|c|c|c|c|}
    \hline
        Platform & \multicolumn{2}{c|}{Nvidia Jetson AGX GPU} & \multicolumn{3}{c|}{Xilinx Zynq UltraScale+ MPSoC} \\
        \hline
        Workflow & TensorFlow & cuDNN & TVM (CPU) & FINN (FPGA) & Vitis-AI (FPGA) \\
        \hline
        Maturity & High & High & High & Low & High \\
        \hline
        Ease of Use & High & Low & High & Low & Medium \\
        \hline
        Documentation and Examples & High & Medium & High & Medium & High \\
        \hline
        Community Support & High & Low & High & Low & High \\
        \hline
    \end{tabular}
\end{table*}

Table \ref{tab:synthesis_engineering} synthesizes the engineering metrics across the five workflows. Overall, TensorFlow and TVM stand out in the comparison. Both are open-source, mature, user-friendly, well-documented, and supported by large, active communities. Close behind, the Vitis-AI workflow exhibits similar positive attributes but is more challenging to use due to its incorporation of proprietary components and a requirement for hardware engineering expertise. Nevertheless, it offers greater customization capabilities than the TVM and TensorFlow workflows. The post-training 8-bit quantization in Vitis-AI, while adding complexity and development time, enhances energy efficiency. The cuDNN workflow is primarily designed for developers of deep neural network frameworks, such as PyTorch and TensorFlow, reflecting its maturity but also its limited suitability for embedded inference. Furthermore, the absence of certain operators, like transposed convolution and nearest neighbor upsampling, necessitates intricate and labor-intensive development. At the bottom of our comparison is the FINN workflow. Its current maturity level is low, with identified bugs, and it presents significant usability challenges. The need to develop custom transformations not available in the FINN library further complicates its usage. Although the community is active, it is relatively small compared to the others. Documentation and examples exist but are dispersed across various websites and GitHub repositories, which complicates the comprehension process. Additionally, FINN’s lack of support for certain operators, such as transposed convolution, necessitates alterations in the neural network architecture.

\subsection{Limitations and Future Works}
\label{sec:limitations_future_works}

The conclusions presented in this paper reflect observations from 2021 to 2023. Nevertheless, the field of neural networks is rapidly evolving, and significant changes in these frameworks are anticipated in the near future. For instance, during the course of our project, we observed maturation in both the FINN and Vitis-AI workflows.

On the GPU front, our research focused on the high-level TensorFlow and the low-level cuDNN workflows. Nvidia's TensorRT, an intermediate, open-source workflow for DNN inference, represents a potential area for future research \cite{NVIDIATensorRT}. Future investigations should also explore quantization to fully leverage the capabilities of Nvidia's Tensor Cores in embedded GPUs, potentially narrowing the energy efficiency gap with FPGAs.

Further research should evaluate the use of more powerful CPUs, such as those based on Intel x86 architectures, with compilers like TVM or Intel nGraph that have shown effectiveness on these processors \cite{li_deep_2020}. Due to time constraints, this study did not explore ASICs for neural network inference, such as Google's Edge TPU or Intel's Movidius VPUs \cite{reuther_survey_2020}, which appear to be promising for embedded applications and warrant future evaluation.

Lastly, the embedded domain poses unique challenges regarding robustness and explainability, aspects not covered in this paper. These topics are currently active research areas in both academic \cite{xu_explainable_2019, nori_interpretml_2019} and industrial spheres \cite{mojsilovic_introducing_2019, ConfianceAI}, deserving attention in future studies.

\section{Conclusion}
\label{sec:conclusion}

This paper has demonstrated the necessity of adapting advanced neural network architectures to novel datasets within an embedded framework. We introduced a lightweight U-Net that achieves the same accuracy with 16 times fewer parameters and Multiply-Accumulate (MAC) operations, validated on an aerial image segmentation dataset \cite{maggiori_can_2017}. Furthermore, this study provided an extensive evaluation and comparison of various methods for real-time semantic segmentation of aerial images, employing three contemporary Commercial Off-The-Shelf (COTS) embedded computers across five distinct workflows.

The FPGA target, utilizing Vitis-AI, emerged as the superior choice due to its performance, energy efficiency, and system maturity. However, its implementation necessitates specialized hardware expertise. The ARM CPU target, leveraging TVM, is notable for its user-friendliness and maturity, yet its relatively low energy efficiency and throughput pose significant challenges for embedded system applications. The GPU target, utilizing TensorFlow, is acknowledged for its maturity and ease of use but is more appropriate for rapid prototyping than for actual embedded solutions. Conversely, the GPU target employing cuDNN is better aligned with embedded deployment but suffers from complexity and a lack of support for various neural network layers. Lastly, the FPGA target using FINN shows high potential for energy-constrained applications but necessitates additional development to become a practical option.

\section{Acknowledgements}

This work was conducted within the SPOC project at the French Institute of Technology (IRT) Saint Exupéry. Funding was provided by the French Research Agency (ANR) and by the industrial partners of the IRT Scientific Cooperation Foundation (FCS).

\bibliography{references}
\bibliographystyle{abbrv}

\end{document}